\documentclass{article}

\usepackage{PRIMEarxiv}

\usepackage[utf8]{inputenc} 
\usepackage[T1]{fontenc}    
\usepackage{hyperref}       
\usepackage{url}            
\usepackage{booktabs}       
\usepackage{amsfonts}       
\usepackage{nicefrac}       
\usepackage{microtype}      
\usepackage{lipsum}
\usepackage{fancyhdr}       
\usepackage{graphicx}       
\usepackage{multirow}
\usepackage{booktabs}
\usepackage{pifont}
\graphicspath{{media/}}     

\title{PC-SAM: Patch-Constrained Fine-Grained Interactive Road Segmentation in High-Resolution Remote Sensing Images}

\author{
  Chengcheng Lv, Rushi Li, Mincheng Wu, Xiufang Shi, Zhenyu Wen, \\
  Zhejiang University of Technology \\
  Hangzhou, China\\
  \texttt{minchengwu@zjut.edu.cn} \\
   \And
  Shibo He \\
  Zhejiang University \\
  Hangzhou, China\\
  \texttt{s18he@zju.edu.cn} \\
}

\begin{document}
\maketitle

\begin{abstract}
  Road masks obtained from remote sensing images effectively support a wide range of downstream tasks. In recent years, most studies have focused on improving the performance of fully automatic segmentation models for this task, achieving significant gains. However, current fully automatic methods are still insufficient for identifying certain challenging road segments and often produce false positive and false negative regions. Moreover, fully automatic segmentation does not support local segmentation of regions of interest or refinement of existing masks.
  Although the SAM model is widely used as an interactive segmentation model and performs well on natural images, it shows poor performance in remote sensing road segmentation and cannot support fine-grained local refinement. To address these limitations, we propose PC-SAM, which integrates fully automatic road segmentation and interactive segmentation within a unified framework. By carefully designing a fine-tuning strategy, the influence of point prompts is constrained to their corresponding patches, overcoming the inability of the original SAM to perform fine local corrections and enabling fine-grained interactive mask refinement.
  Extensive experiments on several representative remote sensing road segmentation datasets demonstrate that, when combined with point prompts, PC-SAM significantly outperforms state-of-the-art fully automatic models in road mask segmentation, while also providing flexible local mask refinement and local road segmentation. The code will be available at https://github.com/Cyber-CCOrange/PC-SAM.
\end{abstract}


\section{Introduction}
With the rapid advancement of artificial intelligence technologies, the recognition and processing of high-resolution remote sensing images have become increasingly efficient, leading to the development of various deep learning–based approaches for remote sensing imagery analysis, including image classification \cite{1}, object detection \cite{2}, semantic segmentation \cite{3}, and change detection \cite{4}. Among the various targets in high-resolution remote sensing images, roads constitute a critical category, serving as the primary infrastructure supporting population mobility, resource circulation, and economic connectivity. Road data plays a vital role in promoting socio-economic development and can be utilized in a wide range of applications, such as traffic flow analysis \cite{5,6}, urban information and service system planning \cite{7,8,9,10}, and disaster prediction and response \cite{11}.

In recent years, most studies on road segmentation in high-resolution remote sensing images have focused on fully automatic methods \cite{12,13,14,15,16}. Although segmentation accuracy has steadily improved, the distribution of remote sensing imagery is significantly more complex than that of natural images, making it challenging to effectively address false positive and false negative regions. Moreover, fully automatic segmentation models often lack flexibility in practical usage: they are unable to selectively segment specific regions of interest, exclude undesired local regions, or refine the segmentation results interactively.

These limitations remain largely unresolved in the field of high-resolution remote sensing road segmentation, motivating us to design a model framework that is both highly flexible and capable of achieving high segmentation accuracy. To retain the advantages of fully automatic approaches, our model supports conventional automatic segmentation while also incorporating point-prompt-based interaction, enabling fine-grained local refinement of the segmentation results. Benefiting from its ability to interpret point prompts and leverage multiple decoders with distinct functionalities, the proposed model can selectively segment regions of interest as well as explicitly exclude undesired regions from segmentation.

\begin{figure}
    \centering
    \includegraphics[width=1.0\linewidth]{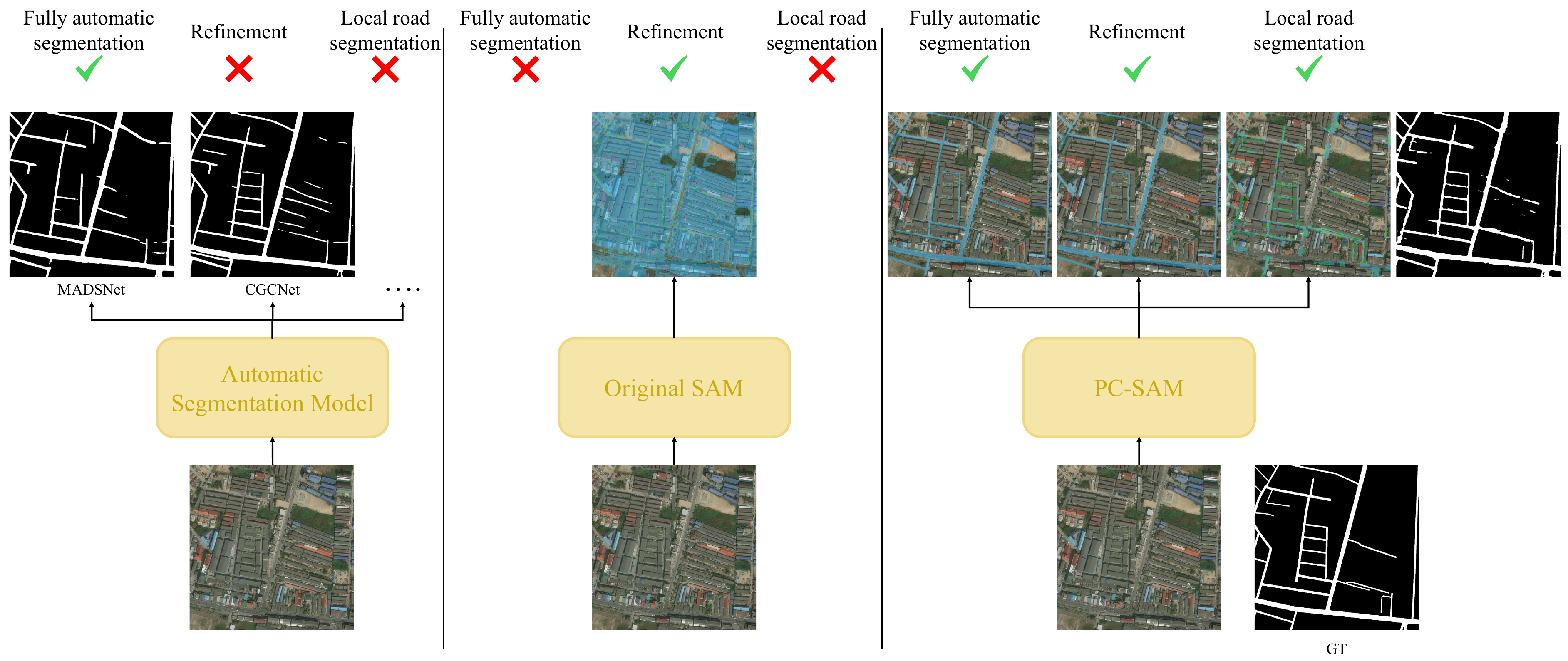}
    \caption{PC-SAM not only integrates fully automatic segmentation and interactive segmentation within a unified framework, but also, by explicitly constraining the refinement range of point prompts to their corresponding patches, enables local road segmentation.}
    \label{fig: PC-SAM Advantage}
\end{figure}

In this paper, we propose PC-SAM, a fine-grained interactive road segmentation model built upon the foundation model \cite{17}. PC-SAM processes positive and negative point prompts using different decoders, and integrates both fully automatic and interactive road segmentation within a unified framework, as shown in Fig.~\ref{fig: PC-SAM Advantage}. Unlike the original interactive foundation model SAM, the effect of point prompts in PC-SAM is strictly constrained, with their influence limited to the corresponding image patch. This design addresses the inherent limitation of SAM, where the influence of point prompts is unpredictable and difficult to control precisely. Such controllability is particularly critical for high-resolution remote sensing road segmentation tasks that require fine-grained refinement.

To simultaneously support both fully automatic segmentation and fine-grained interactive segmentation within a unified framework, we extend SAM by introducing two additional mask decoders, resulting in a model architecture consisting of one image encoder, one prompt encoder, and three mask decoders. Specifically, two of the mask decoders are responsible for automatic road mask decoding with negative point prompts and local road mask decoding with positive point prompts, respectively. The third mask decoder is designed to generate high-recall road masks, which can serve as references in regions where roads are difficult to identify, thereby assisting users in placing positive point prompts more effectively.

In the original SAM model, the influence of point prompts is inherently ambiguous, leading to uncontrollable modifications in the segmentation results. This limitation is unacceptable for high-resolution remote sensing road segmentation tasks that demand precise and fine-grained outputs. To ensure that the influence of point prompts in PC-SAM aligns with the desired behavior and enables accurate local refinement, we design a tailored fine-tuning strategy. During fine-tuning, while adapting SAM’s components to remote sensing imagery, we construct localized supervision signals, including local labels corresponding to positive point prompts and locally missing labels corresponding to negative point prompts. These specially designed labels provide explicit guidance on the effective scope of point prompts, enabling the model to learn well-controlled and interpretable interactions.

Extensive experiments conducted on three widely used public road segmentation datasets demonstrate that, with sufficient point prompts, PC-SAM significantly outperforms previous fully automatic segmentation methods in terms of mask accuracy, while also offering high flexibility in practical usage. Our main contributions can be summarized as follows:
\begin{itemize}
    \item We propose PC-SAM, a flexible and high-precision fine-grained interactive road segmentation model. Within a unified framework, the proposed model integrates both fully automatic segmentation and fine-grained interactive segmentation. To the best of our knowledge, it is the first model in the field of high-resolution remote sensing road segmentation that simultaneously supports fully automatic road segmentation, interactive mask refinement, as well as interactive local inclusion and exclusion of road regions.
    \item To address the limitation of the original SAM, where the influence of point prompts is ambiguous and unsuitable for fine-grained mask refinement, we design a tailored fine-tuning strategy that constrains the effective scope of point prompts. This enables controllable prompt-based refinement, allowing precise and localized corrections of road segmentation masks.
    \item Extensive experiments on three representative public datasets demonstrate that, given sufficient point prompts, PC-SAM achieves superior performance compared to existing fully automatic segmentation methods. These results validate the effectiveness of PC-SAM and highlight its strong potential for road segmentation tasks.
\end{itemize}

\begin{figure*}[htpb]
    \centering
    \includegraphics[width=0.9\linewidth]{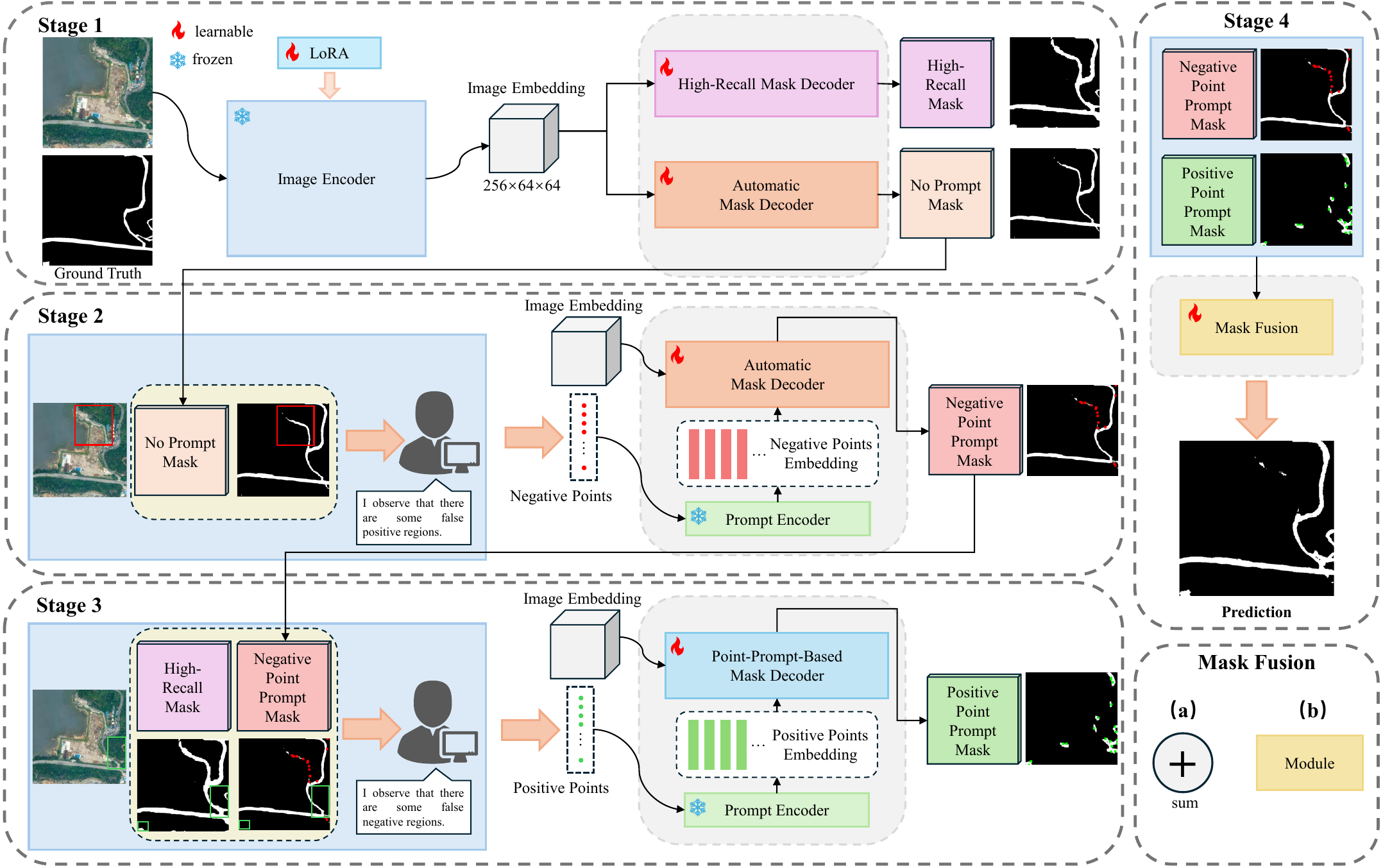}
    \caption{Overall pipeline. Stage 1 performs fully automatic road segmentation, producing both a high-recall road mask and a standard road mask. Stage 2 performs mask removal: based on the image and the fully automatic segmentation mask, users can select regions of no interest and false positive areas using negative point prompts, and the model removes the corresponding road mask regions according to these negative prompts. Stage 3 performs mask addition: based on the high-recall road mask and the mask produced in Stage 2, users can select regions of interest and false negative areas using positive point prompts, and the model segments these regions according to the positive prompts. Finally, the masks from Stage 2 and Stage 3 are fused to obtain the final segmentation result.}
    \label{fig:PC-SAM}
\end{figure*}

\section{Related Work}
\subsection{Foundation Model} 
Recently, new foundation models \cite{20} based on Transformer \cite{18} and Vision Transformer (ViT) \cite{19} have provided powerful and efficient solutions for a wide range of computer vision tasks. Among them, SAM \cite{17} is a widely adopted interactive semantic segmentation model that can achieve high-precision segmentation of arbitrary objects using point prompts, bounding box prompts, and coarse mask prompts. Benefiting from the large-scale dataset used during training and its unique prompt-driven segmentation paradigm, SAM demonstrates strong generalization ability and can effectively adapt to various downstream tasks in a zero-shot manner or with minimal parameter fine-tuning.

\subsection{Conventional Road Segmentation Methods} 
Most existing road segmentation approaches are non-prompt-based and rely on fully automatic segmentation frameworks \cite{21,22,23}. These methods typically enhance the model’s ability to recognize the elongated and continuous structure of roads by incorporating various forms of prior knowledge into the model design or by adopting more advanced architectures. Early approaches, such as D-LinkNet \cite{24}, improve segmentation performance by enlarging the receptive field through dilated convolutions. However, a large receptive field does not necessarily guarantee good topological connectivity of roads. Therefore, subsequent works introduce explicit connectivity modeling and supervision. For instance, CoANet \cite{25} leverages strip convolution tailored to the slender and continuous nature of roads, aiming to suppress background interference during convolution, while also incorporating precise connectivity labels to supervise the model output and improve connectivity. Similarly, OARENet \cite{26} enhances road segmentation performance under dense occlusion by introducing atrous strip convolution. 
Beyond explicit modeling of road priors, improving the intrinsic feature extraction capability of neural networks has also proven effective. For example, MADSNet \cite{27} designs a decoder with strong contextual modeling ability based on a window mechanism, and employs multi-level feature fusion during decoding to enhance segmentation performance. 
Despite these advances, such methods often require complex designs specifically tailored to road structures. Moreover, explicitly modeled priors may not generalize well across different scenarios, and challenging samples still tend to exhibit false positive and false negative regions. In addition, as fully automatic segmentation approaches, these methods lack model-level interactivity and cannot support user-driven refinement of segmentation results.

\subsection{SAM-based Road Segmentation Methods}
SAM has demonstrated strong performance on natural images, and its interactive design introduces new advantages. Several studies have extended SAM to the domain of road segmentation in remote sensing imagery. For instance, Road-SAM \cite{28} adapts SAM to remote sensing data by incorporating high-frequency components of the input images along with lightweight adapters for fine-tuning, enabling accurate fully automatic road segmentation. TPP-SAM \cite{29} exploits trajectory points as foreground prompts and building centroids as background prompts, and further designs a constraint mechanism to select reliable prompts, revealing the zero-shot potential of SAM for road segmentation tasks. GeoSAM \cite{30} fine-tunes SAM using both sparse and dense prompts to achieve automatic segmentation of mobility infrastructure, including roads. RemoteSAM \cite{31} is a powerful vision-language foundation model based on referring expression segmentation, which integrates multiple tasks such as semantic segmentation, object detection, and image classification into a unified framework, and can segment arbitrary geographic targets with high accuracy.

Despite their promising performance, these methods still exhibit notable limitations. Road-SAM, GeoSAM, and TPP-SAM do not support true interactive segmentation. Although RemoteSAM provides strong text-guided interactive segmentation capabilities, it is constrained by the limited granularity of natural language descriptions, making it difficult to precisely control the inclusion or exclusion of specific local road segments in road segmentation tasks.

\section{Methodology}
\subsection{Task Definition}
To achieve flexible and convenient segmentation, the model should be able to perform fully automatic road segmentation, refine local mask predictions based on point prompts, and enable local segmentation within specific regions. We formulate this task as follows: given a remote sensing image $I$ of resolution $H \times W$, the corresponding road mask $M$ also has resolution $H \times W$. We define the local region influenced by point prompts as an image patch of size $l_{h} \times l_{w}$. Accordingly, the mask $M$ can be divided into $\frac{H}{l_{h}} \times \frac{W}{l_{w}}$ non-overlapping patches. Let the patch at row $i$ and column $j$ be denoted as $P_{i,j}$, where $i\in [0, \frac{H}{l_{h}}-1 ]$ and $j\in [0, \frac{W}{l_{w}}-1 ]$. Positive point prompts $PP_{h,w}$ and negative point prompts $NP_{h,w}$ are associated with the patch $P_{\left \lfloor \frac{h}{l_{h}}  \right \rfloor, \left \lfloor \frac{w}{l_{w}}  \right \rfloor }$. We denote this mapping as $g(\cdot)$, defined by
\begin{equation}
\label{eq.gPP}
    g(PP_{h,w})=P_{\left \lfloor \frac{h}{l_{h}}  \right \rfloor, \left \lfloor \frac{w}{l_{w}}  \right \rfloor },
\end{equation}
\begin{equation}
\label{eq.gNP}
    g(NP_{h,w})=P_{\left \lfloor \frac{h}{l_{h}}  \right \rfloor, \left \lfloor \frac{w}{l_{w}}  \right \rfloor }.
\end{equation}

Our goal is to learn a mapping function $f(\cdot)$ that can simultaneously support fully automatic segmentation and point-prompt-based fine-grained local segmentation, such that:
\begin{equation}
\label{eq1}
    f(I)=M,
\end{equation}
\begin{equation}
\label{eq2}
    f(I, \bigcup PP_{h, w})=\bigcup P_{\left \lfloor \frac{h}{l_{h}}  \right \rfloor, \left \lfloor \frac{w}{l_{w}}  \right \rfloor },
\end{equation}
\begin{equation}
\label{eq3}
    f(I, \bigcup NP_{h, w})=M-\bigcup P_{\left \lfloor \frac{h}{l_{h}}  \right \rfloor, \left \lfloor \frac{w}{l_{w}}  \right \rfloor }.
\end{equation}
In practice, we implement this by employing two mask decoders: one is responsible for the mappings in Eq.~\ref{eq1} and Eq.~\ref{eq3}, while the other handles the mapping in Eq.~\ref{eq2}.

\subsection{Patch-Constrained SAM}
As shown in Fig.~\ref{fig:PC-SAM}, the inference process of PC-SAM consists of three stages, integrating fully automatic segmentation with interactive refinement and segmentation. Each module of the model retains the original SAM design, with the only difference being the addition of two extra mask decoders.

\noindent\textbf{Image Encoder.} PC-SAM is built upon the SAM-B architecture and adopts the same pre-trained ViT-B as the image encoder. Consistent with SAM, for each input image, the image encoder is executed only once before the prompt processing. Specifically, the image encoder runs once at Stage 1, and the resulting image embeddings are utilized by the decoders in Stages 1, 2, and 3. Since the pre-trained SAM exhibits limited zero-shot performance on remote sensing road segmentation tasks, we aim to better adapt SAM to the distribution of remote sensing imagery. To this end, we freeze the parameters of the image encoder itself and apply LoRA \cite{32} for parameter-efficient fine-tuning of the image encoder. 

\noindent\textbf{Prompt Encoder.} PC-SAM retains the original prompt encoder from the pre-trained SAM without any modifications and freezes its parameters.

\noindent\textbf{Mask Decoder.} Building on SAM, we introduce two additional mask decoders with identical structures. The High-Recall Mask Decoder (HRMD) provides high-recall road masks as a reference for users, facilitating the placement of positive point prompts. The Automatic Mask Decoder (AMD) performs fully automatic segmentation based on negative point prompts. The Point-Prompt-Based Mask Decoder (PMD) conducts patch-level segmentation guided by positive point prompts. The final segmentation mask is obtained by fusing the output features of AMD and PMD. Formally, AMD implements the mapping functions in Eq.~\ref{eq1} and Eq.~\ref{eq3}, while PMD implements the mapping function in Eq.~\ref{eq2}. PC-SAM inherits the “large encoder–small decoder” architecture of SAM. Consequently, we perform full-parameter fine-tuning only on the lightweight mask decoders, constraining the influence of $PP_{h,w}$ and $NP_{h,w}$ to the patch in which they reside.

\begin{figure}
    \centering
    \includegraphics[width=0.6\linewidth]{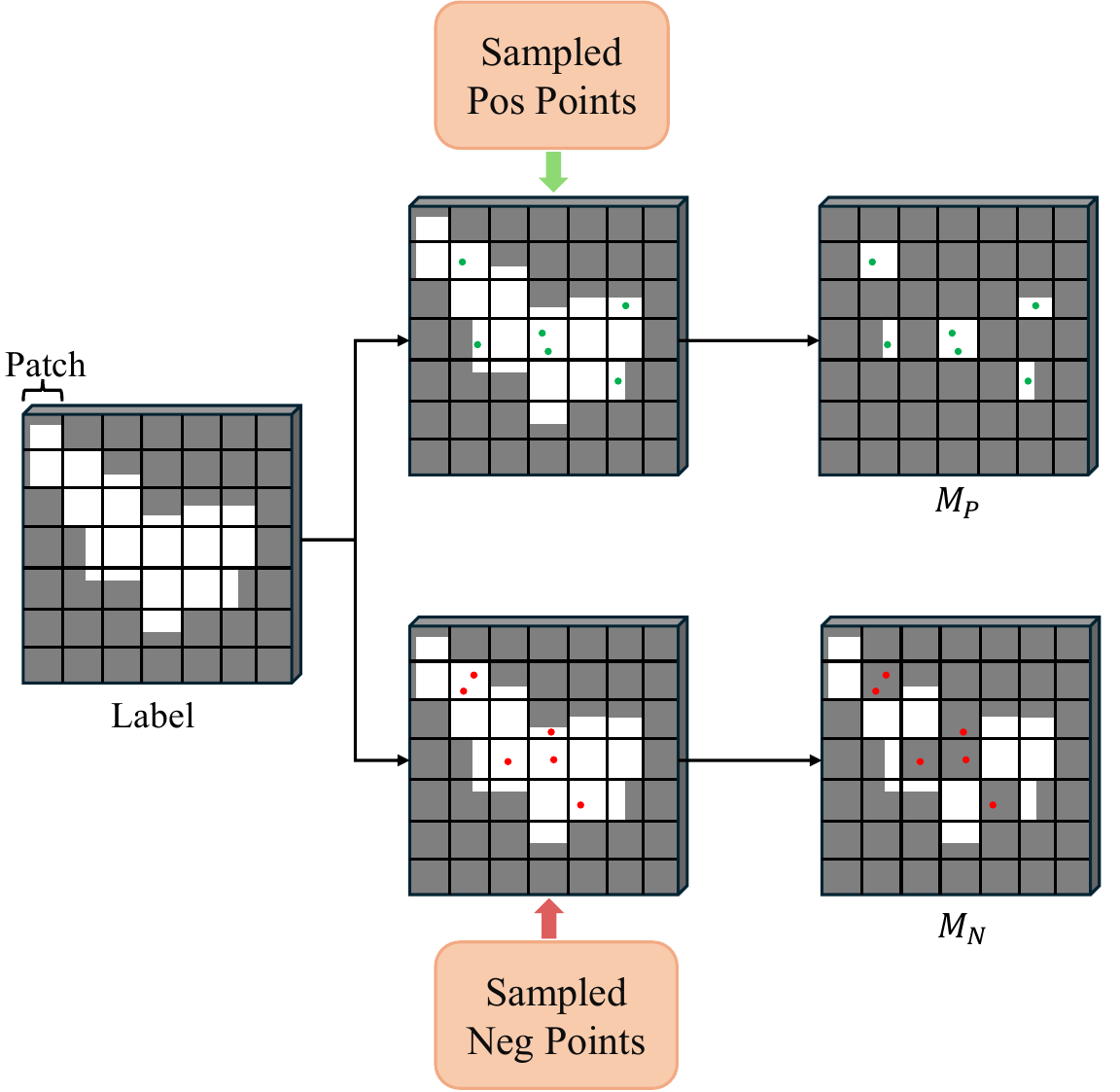}
    \caption{Generation process of $M_{P}$ and $M_{N}$.}
    \label{fig:Masks Generation}
\end{figure}

\subsection{Fine-Tuning Training Strategy}
Since the original SAM cannot control the scope of point-prompt-based modifications, it may produce unintended changes. Therefore, it is necessary to fine-tune SAM’s mechanism for handling point prompts. To constrain the influence of point prompts, we design a fine-tuning training strategy that dynamically generates labels corresponding to randomly sampled point prompts to supervise the model during training.

\noindent\textbf{Random Sampling of Positive and Negative Points.} The random sampling of $PP$ and $NP$ involves both a random quantity and random positions. Let the base total number of sampling points be $N_{B}$ and the proportion of positive points be $R_{F}$. To enable the model to handle varying numbers and ratios of positive and negative prompts, we randomly determine the total number of sampling points $N$ and the positive point ratio $R$ within a specified range for each batch iteration. Accordingly, we introduce a random scaling factor $\delta_{N}$ for the total number of points, and $\delta_{R}$ for the positive point ratio. During training, the total number of sampling points $N$ and the positive point ratio $R$ for the current batch are computed based on $N_{B}, R_{F}, \delta_{N}, \delta_{R}$ by
\begin{equation}
    N=max(0,N_{B}+randint(-N_{B}\times\delta_{N}, N_{B}\times\delta_{N})),
\end{equation}
\begin{equation}
    R=clip(R_{F} + rand_{uniform}(-\delta_{R}, \delta_{R}), 0, 1).
\end{equation}
Here, $randint$ denotes sampling a random integer within a specified range, $rand_{uniform}$ denotes sampling a value from a uniform distribution within a range, and $clip$ limits values to the specified lower and upper bounds if they exceed them. The number of positive point prompts $N_{P}$ and the number of negative point prompts $N_{N}$ are then calculated by
\begin{equation}
    N_{P}=N \times R,
\end{equation}
\begin{equation}
    N_{N}=N - N_{p}.
\end{equation}
After obtaining $N_{P},N_{N}$, we randomly sample $N_{P}$ positive point prompts $PP$ and $N_{N}$ negative point prompts $NP$ from the road foreground regions in the road mask $M$.

\noindent\textbf{Dynamically Generating Point Prompt Label Masks.} The visualization of this process is shown in Fig.~\ref{fig:Masks Generation}. Specifically, positive and negative point prompts are first randomly sampled from the foreground road pixels in the road mask $M$, resulting in a set $PP$ of positive prompts and a set $NP$ of negative prompts, i.e.,
\begin{equation}
    PP=\left \{ PP_{h,w} | (h,w) \in sampled~positive~points \right \},
\end{equation}
\begin{equation}
    NP=\left \{ NP_{h,w} | (h,w) \in sampled~negative~points \right \}.
\end{equation}
The original road mask $M$ is then divided into patches of a pre-defined size $l_{h} \times l_{w}$, and all patches containing any of the sampled points are identified by
\begin{equation}
    P_{P}=\left \{ g(PP_{h,w})|PP_{h,w}\in PP \right \},
\end{equation}
\begin{equation}
    P_{N}=\left \{ g(NP_{h,w})|NP_{h,w}\in NP \right \}.
\end{equation}
Let the label mask corresponding to $PP$ be $M_{P}$. $M_{P}$ retains the original road labels only within the patches $P_{P}$, while all other patches are set to background 0. Formally, this can be expressed by
\begin{equation}
    M_{P}=\bigcup P_{P}.
\end{equation}
The label mask corresponding to $NP$ is denoted as $M_{N}$. $M_{N}$ is obtained by removing from $M$ all patches $P_{N}$. Formally, this can be expressed by
\begin{equation}
    M_{N}=M-\bigcup P_{N}.
\end{equation}

\begin{figure}
    \centering
    \includegraphics[width=1.0\linewidth]{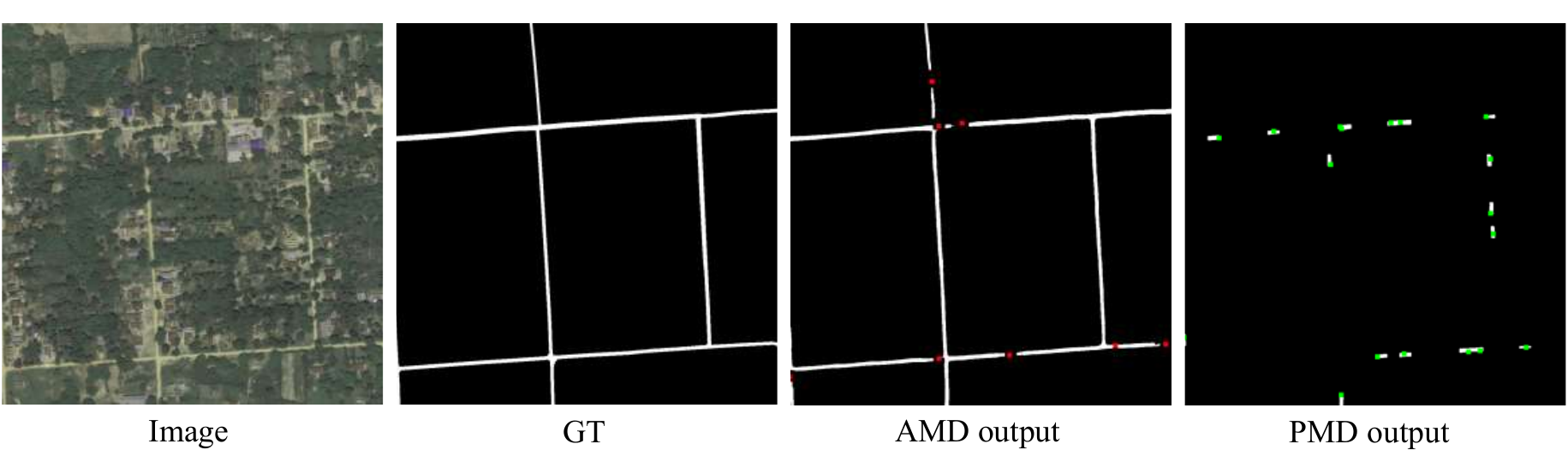}
    \caption{Visualization of the model’s performance. In the figure, red points represent negative prompts, and green points represent positive prompts.}
    \label{fig:train result}
\end{figure}

After obtaining the positive and negative point prompts $PP,NP$ along with their corresponding label maps $M_{P},M_{N}$, the positive prompts $PP$ are fed into the PMD, and the negative prompts $NP$ are fed into the AMD. The outputs of PMD and AMD are supervised using $M_{P}$ and $M_{N}$, respectively. The performance of the trained model is illustrated in Fig.~\ref{fig:train result}.

\begin{figure}
    \centering
    \includegraphics[width=1.0\linewidth]{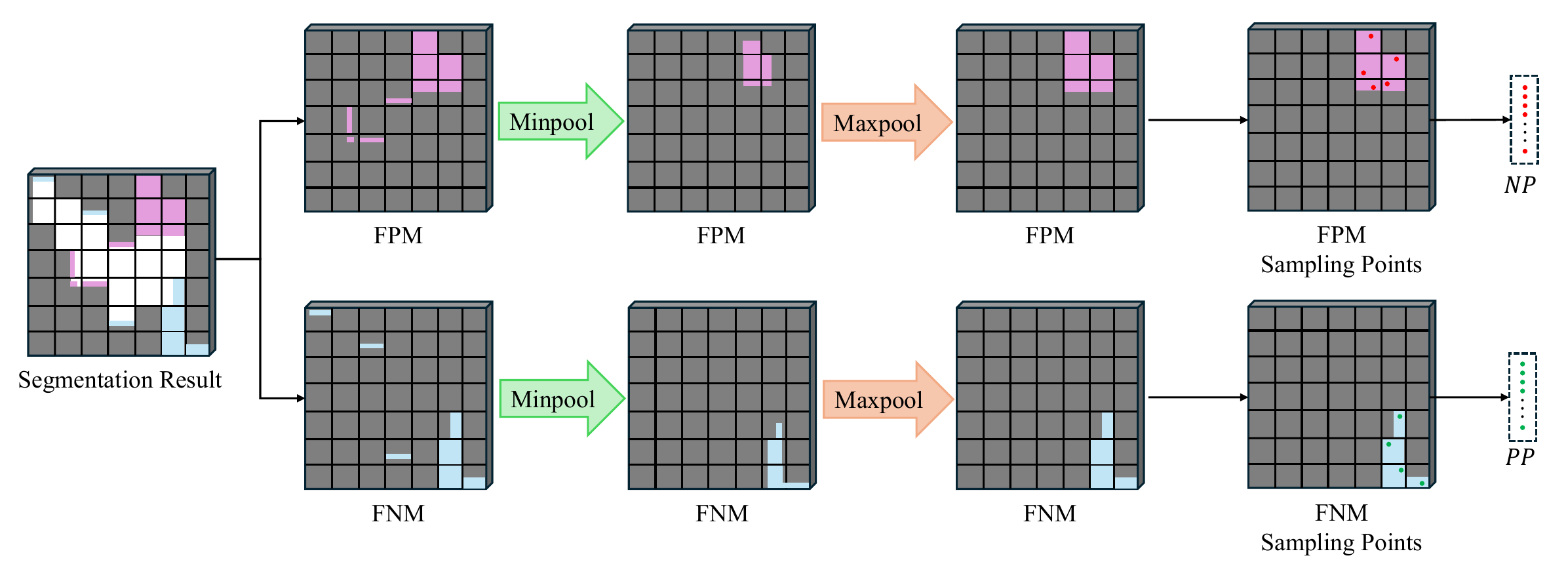}
    \caption{Pipeline of morphological opening and point prompt sampling for segmentation refinement.}
    \label{fig:opening and sampling}
\end{figure}

\subsection{Method for Generating Point Prompts During Testing}
To evaluate the optimal potential performance of PC-SAM, point prompts are required during testing to enable interactive refinement. In manual operation, prompts can be provided gradually based on the segmentation results. However, for automated inference testing, a complete set of point prompts covering all regions that require correction must be generated quickly. To facilitate testing of PC-SAM’s interactive segmentation capability, we design an automated method for generating positive and negative point prompts.

Specifically, as shown in Fig.~\ref{fig:opening and sampling}, point prompts are generated after the Stage 1 fully automatic segmentation. At this point, we have a predicted road segmentation map $S_{A}$ and the ground-truth mask $M$. Based on $S_{A}$ and $M$, we compute the false negative map (FNM) and false positive map (FPM). FNM indicates the road regions that were missed by the segmentation, while FPM identifies regions incorrectly predicted as roads. Positive point prompts are used to correct FNM, and negative point prompts are used to correct FPM.
In practice, there are often many small erroneous pixels along the edges of roads. These tiny points are negligible and do not require correction. Therefore, we apply morphological opening to filter out excessively small regions. To accelerate inference, we implement the erosion and dilation operations in morphological opening using minimum pooling and maximum pooling, which enables efficient parallel computation on the GPU.

After applying morphological opening to obtain FPM and FNM, point prompts are sampled. As in training, FPM and FNM are divided into patches of size $l_h \times l_w$. $NP$ are sampled from foreground regions of FPM patches containing false positives, while $PP$ are sampled from foreground regions of FNM patches containing false negatives.

\begin{table*}[htpb]
  \caption{Quantitative comparison results on the DeepGlobe, Massachusetts Roads and CHN6-CUG datasets.}
  \centering
  \resizebox{\linewidth}{!}{
  \begin{tabular}{c|c|cccc|cccc|cccc}
    \toprule
    & \multirow{2}{*}{Methods} & \multicolumn{4}{c}{DeepGlobe} & \multicolumn{4}{c}{Massachusetts} & \multicolumn{4}{c}{CHN6-CUG}\\ 
    & & Pre.(\%) & Rec.(\%) & IoU(\%) & F1(\%) & Pre.(\%) & Rec.(\%) & IoU(\%) & F1(\%) & Pre.(\%) & Rec.(\%) & IoU(\%) & F1(\%) \\
    \midrule
    \multirow{6}{*}{Automatic Segmentation} & U-Net(MICCAI 2015) \cite{36} & 75.68 & 78.36 & 62.15 & 75.61 & 75.62 & 73.88 & 59.72 & 74.48 & 69.83 & 61.89 & 48.48 & 65.92\\
    
    & D-LinkNet(CVPRW 2018) \cite{24} & 78.10 & 81.69 & 66.21 & 78.89 & 75.58 & 75.16 & 60.51 & 75.14 & 66.71 & 75.90 & 53.90 & 71.84 \\

    & RCFSNet(GRSL 2022) \cite{37} & 81.19 & 78.06 & 65.75 & 78.24 & 69.96 & 80.03 & 59.51 & 74.42 & 77.52 & 62.33 & 52.34 & 65.53 \\

    & OARENet(TGRS 2024) \cite{26} & \textbf{82.12} & 82.26 & 69.98 & 81.58 & \textbf{81.60} & 74.93 & 64.14 & 77.85 & 76.00 & 67.02 & 54.96 & 71.79\\

    & MADSNet(TGRS 2025) \cite{27} & 77.32 & 83.84 & 67.25 & 79.60 & 76.84 & 75.06 & 61.22 & 75.71 & 68.65 & 75.89 & 55.72 & 72.78\\

    & CGCNet(TGRS 2025) \cite{38} & 81.42 & 80.21 & 67.66 & 79.87 & 69.97 & 80.84 & 59.95 & 74.76 & \textbf{78.29} & 71.15 & 58.50 & 71.23 \\

    \midrule
    \multirow{3}{*}{Interactive Segmentation} & PC-SAM(Auto Segment w/o Prompt) & 81.85 & 79.49 & 67.49 & 79.76 & 77.73 & 73.38 & 60.66 & 75.25 & 70.06 & 69.08 & 52.68 & 70.16\\

    & PC-SAM(Auto Segment w/ Prompt) & 80.54 & \textbf{92.04} & \textbf{75.27} & \textbf{85.59} & 75.15 & \textbf{83.14} & \textbf{65.37} & \textbf{78.88} & 72.16 & \textbf{91.61} & \textbf{67.49} & \textbf{81.47}\\
    
    \cmidrule(lr){2-14}
    & Zero-shot SAM-H(ICCV 2023) \cite{17} & 28.55 & 72.15 & 21.72 & 31.54 & 14.74 & 72.59 & 11.83 & 20.12 & 18.99 & 87.31 & 29.10 & 40.87\\
    
    \bottomrule
  \end{tabular}
    }
  \label{tab: Comparision Experiments}
\end{table*}

\subsection{Loss Function}
We adopt Focal Loss \cite{39} and Dice Loss \cite{40} to supervise model training. For the outputs of AMD, PMD, and MFM, the following loss functions are applied, where $S(\cdot)$ denotes the sigmoid function, i.e.,
\begin{equation}
\label{LOSS AMD}
    L_{A}=0.3 \times Dice(S(O_{A}), M_{N}) + 0.7 \times Focal(S(O_{A}), M_{N}),
\end{equation}
\begin{equation}
\label{LOSS PMD}
    L_{P}=0.3 \times Dice(S(O_{P}), M_{P}) + 0.7 \times Focal(S(O_{P}), M_{P}),
\end{equation}
\begin{equation}
\label{LOSS MFM}
    L_{M}=0.3 \times Dice(S(O_{M}), M) + 0.7 \times Focal(S(O_{M}), M).
\end{equation}
When the image size is large while $l_{h}$ and $l_{w}$ are small, the removed regions in $M_{N}$ may become too small. As a result, their contribution to the loss is negligible, making it difficult to optimize through backpropagation. This ultimately leads to the failure of negative point prompts to effectively remove the corresponding patch-level segmentation masks. To address this issue, we introduce an additional loss  $L_{N}$ for optimization. 
First, we take the negative of the AMD's output logits and then apply the sigmoid activation. The resulting probability map is denoted as $p_{A}$. In $p_{A}$, pixels that should correspond to the background will have high probability values. By performing element-wise multiplication of $p_{A}$ and $M$, we extract the probability map of the road regions that should be removed, denoted as $\hat{p_{A}}$. The loss is then computed between $\hat{p_{A}}$ and $(1 - M_N) \times M$. Formally, this can be expressed by
\begin{equation}
    p_{A}=S(-O_{A}),
\end{equation}
\begin{equation}
    \hat{p_{A}}=p_{A} \odot M,
\end{equation}
\begin{equation}
    \hat{M_{N}}=(1-M_{N})\odot M,
\end{equation}
\begin{equation}
\label{LOSS Negative region}
    L_{N}=0.3 \times Dice(\hat{p_{A}}, \hat{M_{N}})+0.7 \times Focal(\hat{p_{A}}, \hat{M_{N}}).
\end{equation}
Notice that $L_{N}$ forces the model to focus on the small regions indicated by negative point prompts that need to be removed, effectively addressing the issue of ineffective negative prompts. 
For the HRMD, to encourage its output to have a high-recall mask, an additional loss $L_{R}$ is introduced that focuses specifically on true positive and false negative regions. Before computing the loss between the output and the label $M$, an element-wise multiplication is performed to filter out false positive regions. This operation biases the optimization of HRMD toward producing masks that cover as much of the road as possible. Formally, this can be expressed by
\begin{equation}
    p_{HR}=S(O_{HR}),
\end{equation}
\begin{equation}
    L_{R} = Loss(p_{HR} \odot M, M),
\end{equation}
\begin{equation}
\label{LOSS HRMD}
    L_{HR}=0.3 \times Dice(p_{HR},M) + 0.65 \times Focal(p_{HR},M) + 0.05 \times L_{R}.
\end{equation}
Combining the losses from AMD (Eq.~\ref{LOSS AMD}), PMD (Eq.~\ref{LOSS PMD}), HRMD (Eq.~\ref{LOSS HRMD}), MFM (Eq.~\ref{LOSS MFM}), and the loss focused on the removed mask regions (Eq.~\ref{LOSS Negative region}), the overall training loss for PC-SAM is defined by
\begin{equation}
\label{LOSS TOTAL}
    L=\alpha_1L_{A}+\alpha_2L_{P}+\alpha_3L_{HRMD}+\alpha_4L_{MFM}+\alpha_5L_{N}.
\end{equation}

\section{Experiments}
\subsection{Experimental Settings}
\noindent\textbf{Datasets.} We conducted experiments on three widely used and representative remote sensing road segmentation datasets: DeepGlobe \cite{33}, Massachusetts Roads \cite{34}, and CHN6-CUG \cite{35}. These datasets cover images of three different sizes and exhibit significant scene variations.
The DeepGlobe dataset contains 6226 images of size 1024×1024 (50cm/pixel), covering urban and rural roads in Thailand, Indonesia, and India. We follow the same data split as in CoANet \cite{25}, which is commonly adopted: 4696 images for training and 1530 images for testing.
The Massachusetts Roads dataset provides 1171 images of size 1500×1500. Using the original split and removing one erroneous image from the training set, we have 1107 images for training and 63 images for testing.
The CHN6-CUG dataset includes 4511 images of size 512×512 (50cm/pixel) from six Chinese cities: Chaoyang District in Beijing, Yangpu District in Shanghai, downtown Wuhan, Nanshan District in Shenzhen, Shatin in Hong Kong, and Macau. The training set contains 3608 images, and the testing set contains 903 images.

\noindent\textbf{Implementation Details.} During training, we set the LoRA parameters with rank $r=8$, and scaling factor $\alpha=32$. The AdamW optimizer is used, with a base learning rate of $lr_{base}=1e-5$ for the AMD and HRMD, and $lr_{base}=1e-4$ for the PMD, MFM and LoRA fine-tuning parameters. A “poly” learning rate schedule is applied, where $lr=lr_{base}\times(1-\frac{iter}{maxiter} ) ^{power}$, with power=3. We train for 200, 200, and 20 epochs on the DeepGlobe, Massachusetts Roads, and CHN6-CUG datasets, respectively. The patch size for point-prompt influence is set to $l_{h} \times l_{w} = 32 \times 32$, and the sampling parameters are $N_B=20,R_F=0.5,\delta_N=1.3,\delta_R=1.0$. Data augmentation includes random flipping, random rotation, and random color jittering. Experiments were conducted on a platform with an Intel Ultra 9 275HX CPU and an NVIDIA RTX 5090 Laptop GPU (24GB).

\subsection{Global Road Segmentation}
We compared the performance of PC-SAM with classical methods such as U-Net \cite{36} and D-LinkNet \cite{24}, as well as several recent state-of-the-art road segmentation methods, including RCFSNet \cite{37}, CGCNet \cite{38}, MADSNet \cite{27}, and OARENet \cite{26}. In addition, we evaluated the zero-shot performance of the pre-trained SAM-H \cite{17} on remote sensing road segmentation. The comparison results are shown in Table \ref{tab: Comparision Experiments}. Compared with recent state-of-the-art fully automatic segmentation methods, PC-SAM achieves superior performance when equipped with point prompts. Specifically, compared with OARENet, PC-SAM improves the IoU by $5.29\%$, $1.23\%$, and $12.53\%$ on the DeepGlobe, Massachusetts, and CHN6-CUG datasets, respectively. This significant improvement stems from the ability of PC-SAM to leverage multiple positive point prompts indicating false negative regions, enabling high-precision segmentation of previously missed local road areas. Meanwhile, negative point prompts allow the model to locally remove falsely predicted road regions (false positives). As a result, PC-SAM substantially improves recall on top of the fully automatic segmentation while maintaining strong precision after prompting.

Existing advanced fully automatic segmentation methods often exhibit inherent limitations due to their model-specific priors, which can be observed from their varying performance across datasets. For example, OARENet incorporates extensive strip convolution with holes, which is particularly effective for capturing thin and occluded road structures, leading to strong performance on the DeepGlobe and Massachusetts datasets. However, on the CHN6-CUG dataset, where roads are generally wider and less occluded, this prior design weakens its contextual modeling capability. CGCNet, benefits from a compact and powerful global context representation, achieving notable performance on the CHN6-CUG dataset. However, this global modeling tends to overlook local details, resulting in a clear performance drop on the Massachusetts dataset, where roads are extremely thin and occupy a very small proportion of the image. MADSNet balances global context modeling with local feature representation, but it does not effectively address the issue of dense occlusions. Consequently, its performance on the DeepGlobe dataset is inferior to other advanced fully automatic segmentation methods.

\subsection{Local Road Refinement}

\begin{table}[htpb]
  \caption{Quantitative performance of PC-SAM for segmenting difficult local road regions missed by automatic segmentation using positive point prompts on the DeepGlobe, Massachusetts Roads, and CHN6-CUG datasets.}
  \centering
  \resizebox{\linewidth}{!}{
  \begin{tabular}{c|cccc}
    \toprule
    Dataset & Pre.(\%) & Rec.(\%) & IoU(\%) & F1(\%) \\
    \midrule
    DeepGlobe & 75.71 & 87.27 & 68.14 & 80.61 \\
    Massachusetts & 52.69 & 60.60 & 39.18 & 56.03 \\
    CHN6-CUG & 72.51 & 80.12 & 61.06 & 75.40 \\
    \bottomrule
  \end{tabular}
    }
  \label{tab: PMD Experiments}
\end{table}

\begin{table}[htpb]
  \caption{Quantitative performance of PC-SAM for removing local erroneous road regions with negative point prompts on the DeepGlobe, Massachusetts Roads, and CHN6-CUG datasets.}
  \centering
  \resizebox{\linewidth}{!}{
  \begin{tabular}{c|c|cccc}
    \toprule
    Dataset & Neg. Prompt & Pre.(\%) & Rec.(\%) & IoU(\%) & F1(\%) \\
    \midrule
    \multirow{2}{*}{DeepGlobe} & \ding{55} & 81.85 & 79.49 & 67.49 & 79.76 \\
    & \ding{52} & 84.05 & 73.85 & 65.34 & 77.78 \\
    \midrule
    \multirow{2}{*}{Massachusetts} & \ding{55} & 77.73 & 73.38 & 60.66 & 75.25 \\
    & \ding{52} & 79.68 & 70.69 & 59.96 & 74.63 \\
    \midrule
    \multirow{2}{*}{CHN6-CUG} & \ding{55} & 70.06 & 69.08 & 52.68 & 70.16 \\
    & \ding{52} & 74.81 & 55.09 & 46.55 & 64.59 \\
    \bottomrule
  \end{tabular}
    }
  \label{tab: AMD Experiments}
\end{table}

Table \ref{tab: PMD Experiments} presents the performance of PC-SAM for local segmentation based on positive point prompts. It can be observed that, on the DeepGlobe and CHN6-CUG datasets, for challenging local road segments that were not captured by PC-SAM's fully automatic segmentation, PC-SAM is able to effectively supplement these segments using positive point prompts. Due to the extremely thin roads and the smaller proportion of road area in the Massachusetts dataset compared to the other two datasets, the performance of PC-SAM in supplementing challenging local segments on Massachusetts is noticeably reduced when using positive point prompts. However, as shown in Table \ref{tab: Comparision Experiments}, despite the decrease in supplementary segmentation performance, the final quality of the model’s mask is still significantly improved.

Table \ref{tab: AMD Experiments} shows the change in mask quality after PC-SAM removes local masks in the fully automatic segmentation that were incorrectly identified as roads using negative point prompts. It can be seen that, across all three datasets, removing erroneous local masks based on negative point prompts significantly improves mask precision. Nevertheless, due to the limitations of the automated negative point prompt generation method, some small areas near correctly segmented masks that do not actually require correction are also sampled with negative points. As a result, when PC-SAM removes these small regions based on negative prompts, it also inadvertently removes part of the correct masks, leading to a decrease in recall and IoU. However, this does not indicate a deficiency of the model itself, as it stems from the inherent limitations of the prompt simulation strategy rather than the model. On the contrary, this further demonstrates that PC-SAM can effectively remove any undesired local masks based on negative point prompts. Overall, according to the results in Table \ref{tab: Comparision Experiments} for PC-SAM under fully automatic segmentation and with all point prompts, both positive and negative point prompts have a positive effect on improving the final mask quality.

\subsection{High-Recall Mask}

\begin{table}[htpb]
  \caption{Quality of high-recall masks across different datasets.}
  \centering
  \resizebox{\linewidth}{!}{
  \begin{tabular}{c|cccc}
    \toprule
    Dataset & Pre.(\%) & Rec.(\%) & IoU(\%) & F1(\%) \\
    \midrule
    DeepGlobe & 39.73 & 97.86 & 39.00 & 53.38 \\
    Massachusetts & 47.98 & 91.31 & 45.17 & 61.28 \\
    CHN6-CUG & 38.62 & 93.22 & 37.01 & 53.77 \\
    \bottomrule
  \end{tabular}
    }
  \label{tab: HRMD Experiments}
\end{table}

PC-SAM can provide high-recall road masks for users as a reference when selecting positive point prompts. The quality of these masks on each dataset is shown in Table \ref{tab: HRMD Experiments}. It can be observed that the masks produced by HRMD achieve a high recall while maintaining a certain level of quality.

\subsection{Visualization}
\begin{figure}
    \centering
    \includegraphics[width=1.0\linewidth]{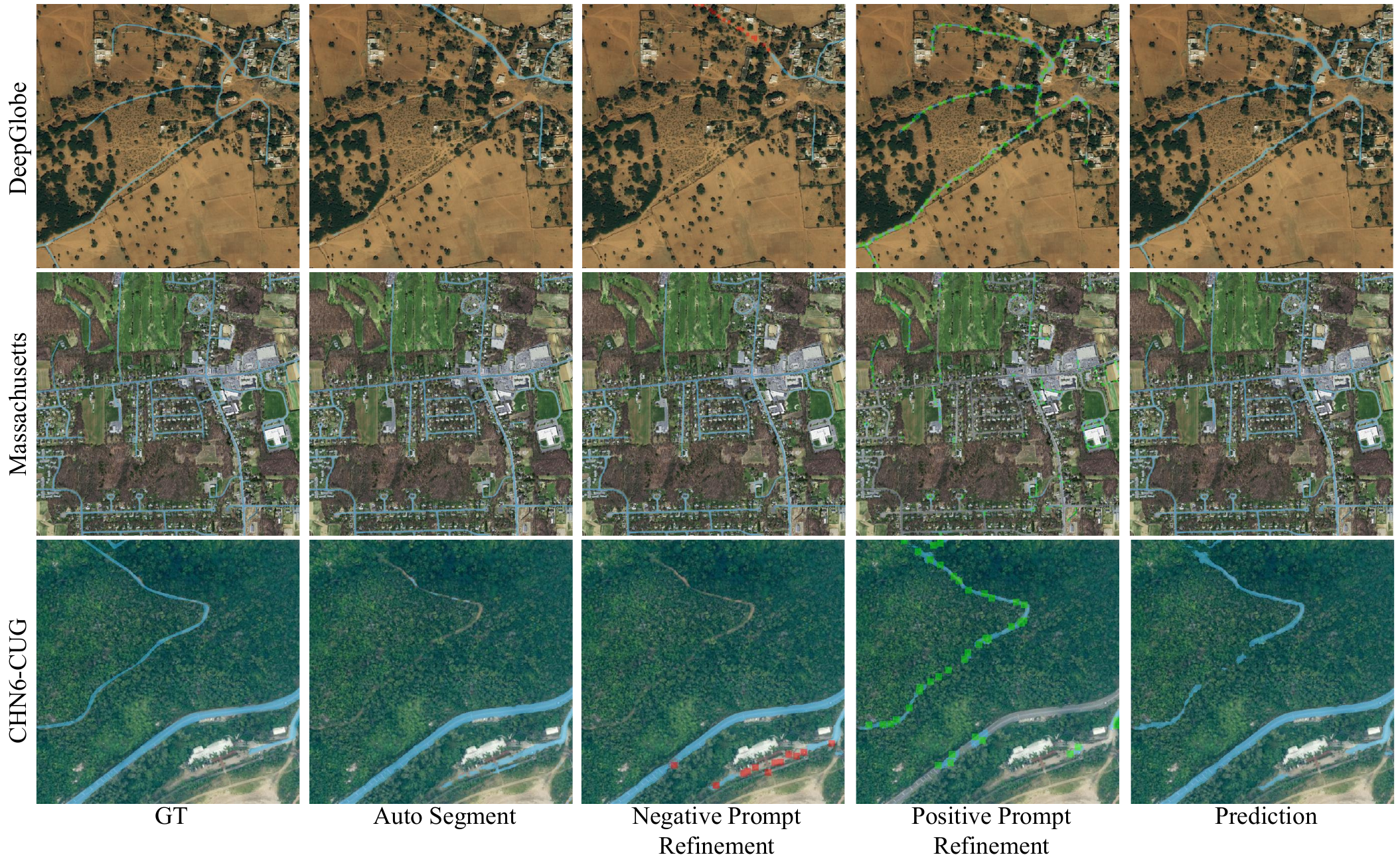}
    \caption{Visualization of the segmentation process of PC-SAM across different datasets, where red points denote negative point prompts and green points denote positive point prompts.}
    \label{fig: PC-SAM Segmentation}
\end{figure}

Fig.~\ref{fig: PC-SAM Segmentation} illustrates the visualization of PC-SAM across three datasets, including the progression from fully automatic segmentation to prompt refinement and the final mask. It can be observed that, through multiple point prompts, PC-SAM effectively achieves both local and large-scale corrections of road masks. This capability cannot be achieved by either the original SAM or fully automatic segmentation, highlighting a unique advantage of PC-SAM.

\begin{figure*}[htpb]
    \centering
    \includegraphics[width=1.0\linewidth]{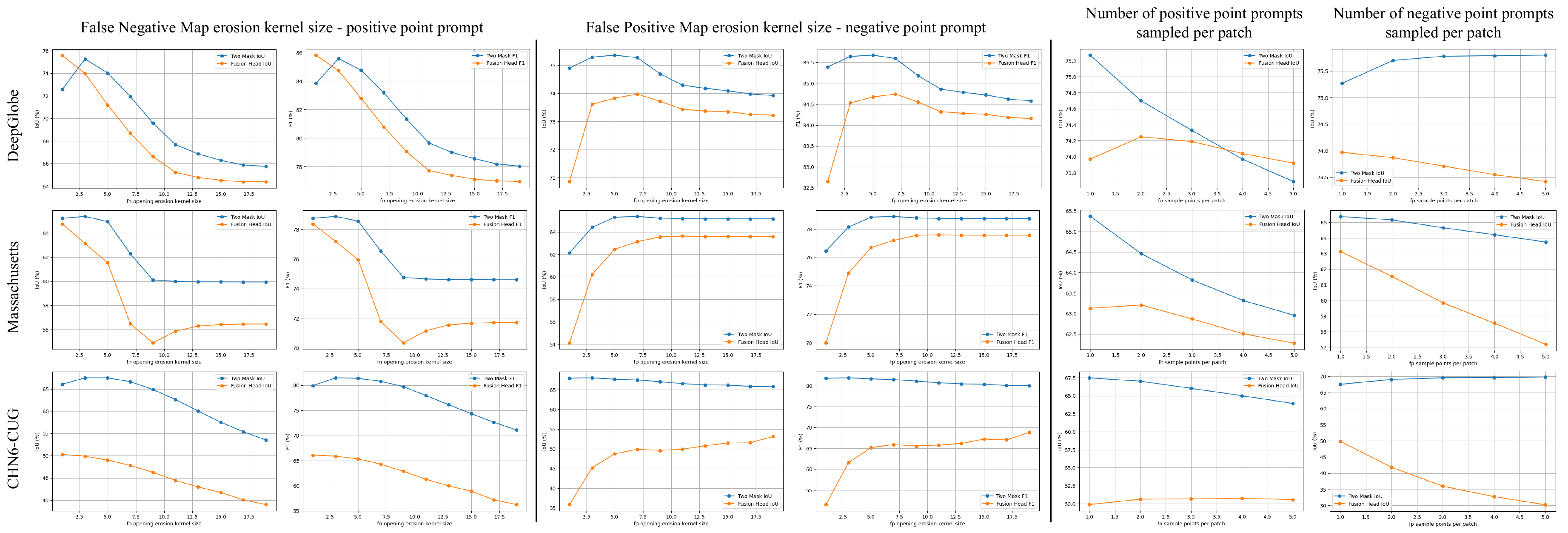}
    \caption{Ablation studies on the erosion kernel size in the morphological opening operation and the point prompt sampling density.}
    \label{fig: Kernel size and sampling number ablation}
\end{figure*}

\subsection{Ablation Study}

\begin{table}[htpb]
  \caption{Experimental results of different mask fusion strategies.}
  \centering
  \resizebox{\linewidth}{!}{
  \begin{tabular}{c|c|cccc}
    \toprule
    Dataset & Strategy & Pre.(\%) & Rec.(\%) & IoU(\%) & F1(\%) \\
    \midrule
    \multirow{2}{*}{DeepGlobe} & Sum & 80.54 & 92.04 & 75.27 & 85.59 \\
    & Fusion & 85.58 & 84.47 & 73.97 & 84.74 \\
    \midrule
    \multirow{2}{*}{Massachusetts} & Sum & 75.15 & 83.14 & 65.37 & 78.88 \\
    & Fusion & 81.91 & 73.13 & 63.13 & 77.20 \\
    \midrule
    \multirow{2}{*}{CHN6-CUG} & Sum & 72.16 & 91.61 & 67.49 & 81.47 \\
    & Fusion & 80.14 & 57.13 & 49.88 & 65.88 \\
    \bottomrule
  \end{tabular}
    }
  \label{tab: Mask Fusion Experiments}
\end{table}

\noindent\textbf{Mask Feature Fusion Strategy.} PC-SAM offers two mask fusion strategies: (1) Direct Mask Merging, which combines the output masks of AMD and PMD, and (2) the Mask Fusion Module (MFM), which fuses the decoders’ output features. Experiments on these strategies are shown in Table \ref{tab: Mask Fusion Experiments}. Across all datasets, direct mask merging performs better, as regions indicated by positive and negative point prompts are generally correct, making additional fusion unnecessary. Nevertheless, MFM improves mask precision, which can be advantageous when precision is critical. Therefore, PC-SAM retains MFM while outputting masks for both strategies.

\noindent\textbf{Level of Refinement.} 
The granularity of point-prompt-based refinement corresponds to the size of the region subject to correction. It depends on the visibility of the area and is related to the cost of mask refinement. Therefore, we conducted an ablation study on the kernel size of the erosion operation in the morphological opening used in the automated point sampling method during testing. A larger erosion kernel increases the area of erroneous mask that must be covered to trigger point-prompt-based correction. Additionally, we performed an ablation study on the density of point prompts sampled per patch. The results are shown in Fig.~\ref{fig: Kernel size and sampling number ablation}. 
For positive point prompts, sampling every false-negative pixel does not necessarily yield the optimal mask. As shown in Fig.~\ref{fig: Kernel size and sampling number ablation}, for the FNM erosion kernel ablation study, when the erosion kernel size is 1, the mask quality is lower than that obtained with a kernel size of 3. For instance, when a few pixels along the sides of a road are missing in the mask, using positive point prompts can actually introduce a large number of erroneous mask regions. In practice, missing masks for individual pixels are generally imperceptible. Therefore, we report the optimal results using an FNM erosion kernel of size 3.
For negative point prompts, the model achieves the best overall performance when the FPM erosion kernel size is set to 7, and the optimal results we report are based on this setting. Fig.~\ref{fig: Kernel size and sampling number ablation} also presents the ablation study on the density of point prompts. The results indicate that, for positive point prompts, sampling just one per patch is sufficient to achieve optimal performance, and denser positive prompts can actually reduce mask quality, which aligns with our training strategy. For negative point prompts, although higher sampling density may slightly improve mask quality, this is mainly because a large number of regions from fully automatic segmentation are removed, followed by extensive use of positive point prompts for refinement. However, this significantly increases the cost of positive point prompting. Therefore, we also report the optimal model performance using only one negative point prompt sampled per patch.

\section{Conclusion}
In this paper, we propose PC-SAM, a novel SAM-based fine-grained interactive road segmentation model for remote sensing images. By constraining the refinement range of point prompts, PC-SAM integrates fully automatic road segmentation, fine-grained interactive mask refinement, and fine-grained interactive local road segmentation within a unified framework. Extensive evaluation experiments demonstrate that, compared to current state-of-the-art automatic segmentation models, PC-SAM offers the advantage of flexible segmentation and, when provided with point prompts, achieves a significant improvement in mask quality. PC-SAM provides a practical, high-performance, and user-friendly solution for road segmentation in remote sensing images.


\end{document}